\pdfoutput=1

\documentclass[11pt]{article}

\usepackage{acl}

\usepackage{times}
\usepackage{latexsym}

\usepackage{subfig}
\usepackage{graphicx}

\usepackage[T1]{fontenc}

\usepackage[utf8]{inputenc}

\usepackage{microtype}

\usepackage{tikz}
\usetikzlibrary{bayesnet}
\usepackage{algorithm}
\usepackage{amsmath}
\usepackage{amssymb}
\usepackage{bm}

\newcommand{\bX}{\bm{X}}

\newcommand{\bW}{\bm{W}}

\newcommand{\todo}[1]{\textbf{\color{red}[TODO #1]}}

\usepackage{comment}
\newenvironment{removed}{}{}
\excludecomment{removed}

\title{Conceptualizing Treatment Leakage in Text-based Causal Inference}

\author{Adel Daoud\textsuperscript{\normalfont1,3,4}\hspace{6mm}Connor T. Jerzak\textsuperscript{\normalfont1,3}\hspace{6mm}Richard Johansson\textsuperscript{\normalfont2,4} \\
  \textsuperscript{1}Linköping University, 
  \textsuperscript{2}University of Gothenburg,\\
  \textsuperscript{3}Harvard University, \textsuperscript{4}Chalmers University of Technology\\
  \texttt{\{adel.daoud, connor.jerzak\}@liu.se, richard.johansson@gu.se} \\}

\begin{document}
\maketitle
\begin{abstract}
Causal inference methods that control for text-based confounders are becoming increasingly important in the social sciences and other disciplines where text is readily available. However, these methods rely on a critical assumption that there is no \emph{treatment leakage}: that is, the text only contains information about the confounder and no information about treatment assignment. 
When this assumption does not hold, methods that control for text to adjust for confounders 
face the problem of post-treatment (collider) bias. However, the assumption that there is no treatment leakage may be unrealistic in real-world situations involving text, as human language is rich and flexible. Language appearing in a public policy document or health records may refer to the future and the past simultaneously, and thereby reveal information about the treatment assignment.

In this article, we define the treatment-leakage problem, and discuss the identification as well as the estimation challenges it raises. Second, we delineate the conditions under which leakage can be addressed by removing the treatment-related signal from the text in a pre-processing step we define as \emph{text distillation}. Lastly, using simulation, we show how treatment leakage introduces a bias in estimates of the average treatment effect (ATE) and how text distillation can mitigate this bias.
\end{abstract}

\section{Introduction}
In observational settings, scholars need to collect information about potential confounders in order to estimate the causal effect ($\tau$) of a treatment on an outcome \cite{daoudStatisticalModelingThree2020}. If we observed the set of confounders directly, we could condition on those quantities to recover unbiased causal effects. Yet, because some confounders $U$ are difficult to measure directly, scholars are turning to alternative data sources, such as medical records, policy documents, or social media posts, to indirectly measure (proxy) confounders \cite{kinoScopingReviewUse2021}. Recent methodological frameworks supply ways of integrating  high-dimensional text data into causal estimation \citep{mozer2020,roberts2020,feder2021}. 

However, prior literature has primarily assumed that documents only contain information about the confounder, but not about the treatment---something we term the \emph{no-treatment-leakage assumption}. Here, ``contain information'' means that the text is caused by the treatment (or the confounder) directly or indirectly.  When  treatment leakage occurs after treatment assignment, its bias is equivalent to a post-treatment bias \cite{Pearl2015}.

Treatment leakage leads to an identification challenge. The challenge is that $\bW$ is both necessary for adjusting (as it is a proxy) yet it is also a post-treatment variable. Without treatment leakage, $\bW$ would not be a post-treatment variable, as it does not harbour information about the treatment assignment. But because of leakage, scholars would have to accept bias arising from either adjusting on a post-treatment variable (arising from the part of $\bW$ influenced by the treatment) or bias arising from not adjusting for unobserved confounding. Although several methodological studies develop and adapt causal-inference methods for text data \cite{keith2020}, almost no studies examine the biasing influence of treatment leakage and how to counter this bias. 

Our work investigates the treatment-leakage challenge. It shows that if $\bW$ is the only available text representing $U$ and there exists a distillation method, $f$, that has the ability to transform (e.g. partition) $\bW$ into its post-treatment $\bW_T$ and proxy  textual-components $\bW_U$, then adjusting on $\bW_U$ is the best one can do in identifying $\tau$. As $\bW_U$ is not post-treatment, we can adjust for it to reduce the bias when estimating $\tau$. These $f$ functions can represent a human annotator, identifying and removing parts of text (e.g., words, sentences) that belong to $\bW_T$ and curating $\bW_U$; or, under additional assumptions, $f$ can be based on supervised or unsupervised machine learning models that transform the text or its representation \cite{akerstromNaturalLanguageProcessing2019,feder2021}. 

In this paper, we define key assumptions and demonstrate the mechanics of text distillation in a simulated experiment. Using a language model, 
we generate synthetic documents $\bW$  
so that they contain information about the
treatment assignment, $T$, and the unobserved confounding, $U$, imprinted paragraph by paragraph. Because we control which paragraph is affected by $T$ (injecting post-treatment bias) or by $U$ (infusing knowledge about the confounder), we have an oracle distillation function, $f$, that mimics human coding. This oracle method perfectly distills $\bW$, and supplies $\bW_U$. Then, when using $\bW_U$ in our causal model, we reduce bias of $\hat{\tau}$ markedly. Although our oracle is idealized, it deepens intuition, and in future work, we will investigate the conditions under which automated methods can be applied to obtain $f$. 

By conceptualizing the problem of treatment leakage in text data and investigating its impact, 
scholars developing causal methods can be better positioned to tailor their frameworks to reduce bias; domain scholars can better calibrate their data collection procedure to account for this leakage.

\section{Treatment Leakage in Text Data}
While the literature on dealing with confounding in observational studies is established \cite{Rubin1974}, recent advances have been made in the analysis of text-based causal inference. Indeed, text $\bW$ is widely available in the health and social sciences \citep{gentzkow2019text, kinoScopingReviewUse2021}, and can be used to proxy for some confounders, $U$, that would otherwise remain unobserved \cite{keith2020}. If the text only contains information about $U$ and no other factors, then $\bW$ is a faithful representation of $U$ and we denote it as $\bW_U$. 
However, text, by its nature as a medium of creativity, rarely has fixed boundaries, and can contain information not only about confounders, but also leak information about the treatment assignment and its effects. 



The future- and backward-looking nature of text can exacerbate treatment leakage. Documents that often contain backward looking temporally (e.g. in much of journalism) or has an unknown production date, will like contain information about the treatment and its effects. Using these documents directly for causal inference would inject post-treatment bias. Conversely, documents that reference the future (e.g., many public-policy documents in the economy and polity) may also lead to unfavorable RMSE if they predict the future well (see \S\ref{ss:otherCases}). As a result, a substantial amount of real-world text containing rich information about confounding factors might be affected by that language can reference the future, post-treatment state.



\subsection{Characterizing Treatment Leakage}
\label{ss:bias}

We define \emph{treatment leakage} as when  the text, $\bW$, is affected by treatment status, $T$: that is, $\bW$ is conditionally dependent on $T$ given $U$.
\[
\textrm{ \emph{Treatment leakage:} } \; \bW \not\perp T \, | \, U
\]
The treatment leakage can take different forms. In the most straightforward case, we can assume that a portion $\bW_T$ is affected by $T$ while another portion $\bW_U$ is affected by $U$. However, in the general case it may be difficult to partition the document into treatment- and confounder-related passages, and we should see $T$ and $U$ as latent factors controlling the data-generating process. For instance, $T$ may affect the overall tone or sentiment of a document.

We can quantify the degree of treatment leakage in different ways. If the text can be partitioned into treatment- and confounder-related passages $\bW_T$ and $\bW_U$ as described above, we can consider the fraction $\frac{|\bW_T|}{|\bW|}$ to be a measure of the degree of treatment leakage; this also assumes that each partition carries strength equal to the number of its elements (e.g., words) and each element has the same strengths. In the general case, we may turn to information-theoretical quantities, for instance the conditional mutual information between $\bW$ and $T$ given $U$.

In the following, we discuss a number of situations in which treatment leakage can occur.

\subsubsection{Case 1: Text is Post-treatment}\label{ss:case1}
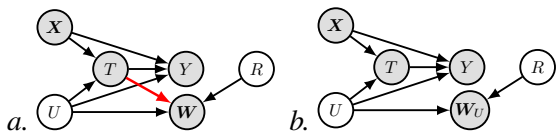
\begin{figure}[htb]
    \begin{center}
\subfloat{\emph{a.}}{
    \begin{tikzpicture}[line width=0.25mm, scale=0.65, transform shape]
        \node[obs] (X) {$\bX$} ; %
        \node[latent, below=of X] (U) {$U$} ; %
        \node[obs, right=of X, xshift=-6mm, yshift=-8.5mm] (T) {$T$} ; %
        \node[obs, right=of T, xshift=-2mm] (Y) {$Y$} ; %
        \node[latent, right=of Y, xshift=-3mm, yshift=0mm] (R) {$R$} ; %
        \node[obs, right=of U, xshift=9.5mm] (W) {$\bW$} ; %
        \edge[-latex] {X,U} {T} ; %
        \edge[-latex] {X,U,T} {Y} ; %
        \edge[-latex] {U} {W} ; %
        \edge[-latex] {R} {W} ; %
        \edge[-latex, line width=0.33mm, red] {T} {W} ; %
    \end{tikzpicture}
}
\subfloat{\emph{b.}}{
    \begin{tikzpicture}[line width=0.25mm, scale=0.65, transform shape]
        \node[obs] (X) {$\bX$} ; %
        \node[latent, below=of X] (U) {$U$} ; %
        \node[obs, right=of X, xshift=-6mm, yshift=-8.5mm] (T) {$T$} ; %
        \node[obs, right=of T, xshift=-2mm] (Y) {$Y$} ; %
        \node[latent, right=of Y, xshift=-3mm, yshift=0mm] (R) {$R$} ; %
        \node[obs, right=of U, xshift=9.5mm] (W) {$\bW_U$} ; %
        \edge[-latex] {X,U} {T} ; %
        \edge[-latex] {X,U,T} {Y} ; %
        \edge[-latex] {U} {W} ; %
        \edge[-latex] {R} {W} ;
    \end{tikzpicture}
}
\end{center}
\caption{A causal model consisting of observed variables (shaded): confounders ($\bX$), treatment ($T$),  outcome ($Y$), document ($\bW$), and unobserved variables (unshaded):  confounder ($U$) and residual factors ($R$). The red-colored edge in \emph{a.} represents the treatment leakage. In \emph{b.}, A distillation function $f$ has removed the treatment information in the text, leaving only information from the confounder. A perfect intervention of $f$ is equivalent with deleting the red arrow; a less than perfect intervention reduces at least its dependence. \label{fig:causalmodel}}
\end{figure}

In one form of this phenomenon, there is a causal relationship between the treatment status $T$ and the text.
Figure \ref{fig:causalmodel}, panel \emph{a.}, shows a directed acyclic graph (DAG) representing this scenario where the text affected by the treatment status. 
This sort of treatment leakage induces  post-treatment bias: when the text is affected by the treatment, conditioning on the text (which is a collider) opens the path from $T$ to $Y$ through $\bW$ and $U$, will in general yield biased estimates (in the notation of \citet{Pearl2015}, $(Y \not\perp T | \bW)_{G_{\underline{T}}}$). 

Identification assumptions may also be hard to maintain, with the treated/control units having distinct text features (e.g. if all treated units have associated texts referring to the treatment). This lack of overlap 
would violate the identification assumptions of causal estimators such as Inverse Propensity Score Weighting (IPW) \citep{heinrich2010primer}, and could lead to extreme estimated probabilities, something we see empirically in Figure \ref{fig:histogram}.

\subsubsection{Other Cases}\label{ss:otherCases}
Figure \ref{fig:causalmodel} shows a case when text is post-treatment,  but in other cases the precise DAG structure may not be known. For example, text may represent a mediator if the document includes post-treatment information and also affected the outcome (if, for example, the text is congressional speech and the outcome is a roll call vote). If the proxy text is pre-treatment and directly affects the treatment, conditioning on the treatment-related portion of the text could increase the variance of estimation, leading to unfavorable RMSE \citep{myers2011effects}. 

\section{Text Distillation as Preprocessing}
\emph{Text distillation} is a form of text preprocessing. It has to target any text (e.g., tone, words, sentences) that belongs to $\bW_T$, and remove it from $\bW$. Thus, distillation ensures that the treatment signal is negated. As Figure \ref{fig:causalmodel}, panel \emph{b.} shows, if distillation is perfectly successful, it results in cutting the red arrow (from $T$ to $\bW$). The arrow is cut, because the distillation function has removed  $\bW_T$ from $\bW$, supplying $\bW_U$ for causal analysis.

\subsection{Assumptions for Valid Distillation}

Depending on how the treatment leakage is manifested in $\bW$, we need to introduce assumptions to make distillation feasible. As already discussed in $\S\ref{ss:bias}$, in some cases we may assume that $\bW$ contains treatment-related passages $\bW_T$ and confounder-related passages $\bW_U$. We may further assume that the text is \emph{separable}: that is, $\bW_T$ and $\bW_U$ do not overlap.
\[
\textrm{ \emph{Separability Assumption:} } \;\; \bW_U \cap \bW_T = \varnothing
\]
Assuming separability, a perfect distillator will produce $\bW^* = f(\bW)$ that is equivalent to the confounder-related portion of the text, $\bW_U$. \emph{Perfect} distillation means that the distillator $f$ identified text that contains the same information about $U$ as $\bW_U$ has. Thus, if $\bW_U$ is a valid adjustment set, then $\bW^*$ is that as well. The separability assumption is appealing because it implies that researchers only need to find a valid partition of the text (and do not need to consider all possible text transformations). 




This separability assumption is particularly plausible for text data, which by its nature consists of a sequence of linguistic signifiers which can be decomposed into smaller units (e.g. paragraphs). 

While plausible for many circumstances, in some cases separability may not hold, as when the entire tone of the text is affected by the treatment. In this more complicated setting, we need a more general assumption, that the transformed text, $\bW^*$, is conditionally independent of $T$ given $U$. That is, the conditional mutual information between $\bW^*$ and $T$ given $U$ is zero, while information about $U$ in $\bW^*$ is maintained. Despite the benefits of this more general framing, because $U$ is unobserved, it may be difficult for investigators to
assess whether the assumption is satisfied or whether ethically problematic information has been included in the $f$ function (e.g., race; \citet{menon2018cost}). Unlike numerical data, as text data is readable, scholars can examine and validate whether $\bW^*$ still contains information about $T$.

\section{Experimental Setup}
\label{sec:method}


We use simulation to illustrate the dynamics of text distillation and build on the framework for evaluating text-based causal inference methods introduced by \newcite{wooddoughty2021}. 
We generate numerical covariates from the model in Figure~\ref{fig:causalmodel}; the general procedure is described in \S\ref{app:general}, with implementation details in \S\ref{app:specific}. Parameters are selected so that ATE estimates $\hat{\tau}$ are biased if the estimator does not account for the unobserved confounder $U$.

Following \newcite{wooddoughty2021}, we generate documents, $\bW$, by sampling from an English-language GPT-2 model \cite{radford2019}. 
In contrast to their approach, text generation is conditioned not only on $U$ but also on $T$. As described in detail in \S\ref{app:general}, we define paragraph-level topics, where some topics are associated with $U$, some with $T$, and some with a 
residual topic 
related only to other background variables ($R$ in Figure~\ref{fig:causalmodel}).
For a given paragraph topic, we define a number of prompts and a distribution shift that increases the probability of generating topic-related keywords.

As we simulate and record which paragraphs are affected by $T$ and by $U$, our distillator $f$ has oracle properties. We can then use $f$ to investigate three idealized distillation scenarios. The first is when a distillator was not applied or the distillator failed to do any distillation $f(\bW)=\bW$. It outputs the same corpus.
The second is when it perfectly distills $\bW$, excluding all paragraphs affected by  $T$. That is, apply $f(\bW)=\bW^*$ such that $\bW^*=\bW_U$. The third scenario is when $f$ was overly aggressive and accidentally removed not only paragraphs related to $T$ but also those related to $U$, resulting in $\bW^{**}$. This corpus violates the proxy-faithfulness assumption that $\bW^{**}$ fully measures $U$. Then, we use the three corpora, one at a time, for causal inference. We use an Inverse Propensity Weighting (IPW) estimator, fully described in Appendix \ref{sec:ipwdetails}.


\section{Experiments and Results}\label{sec:experiments} 

\begin{figure}[htb]
 \begin{center}  \includegraphics[width=0.6\linewidth]{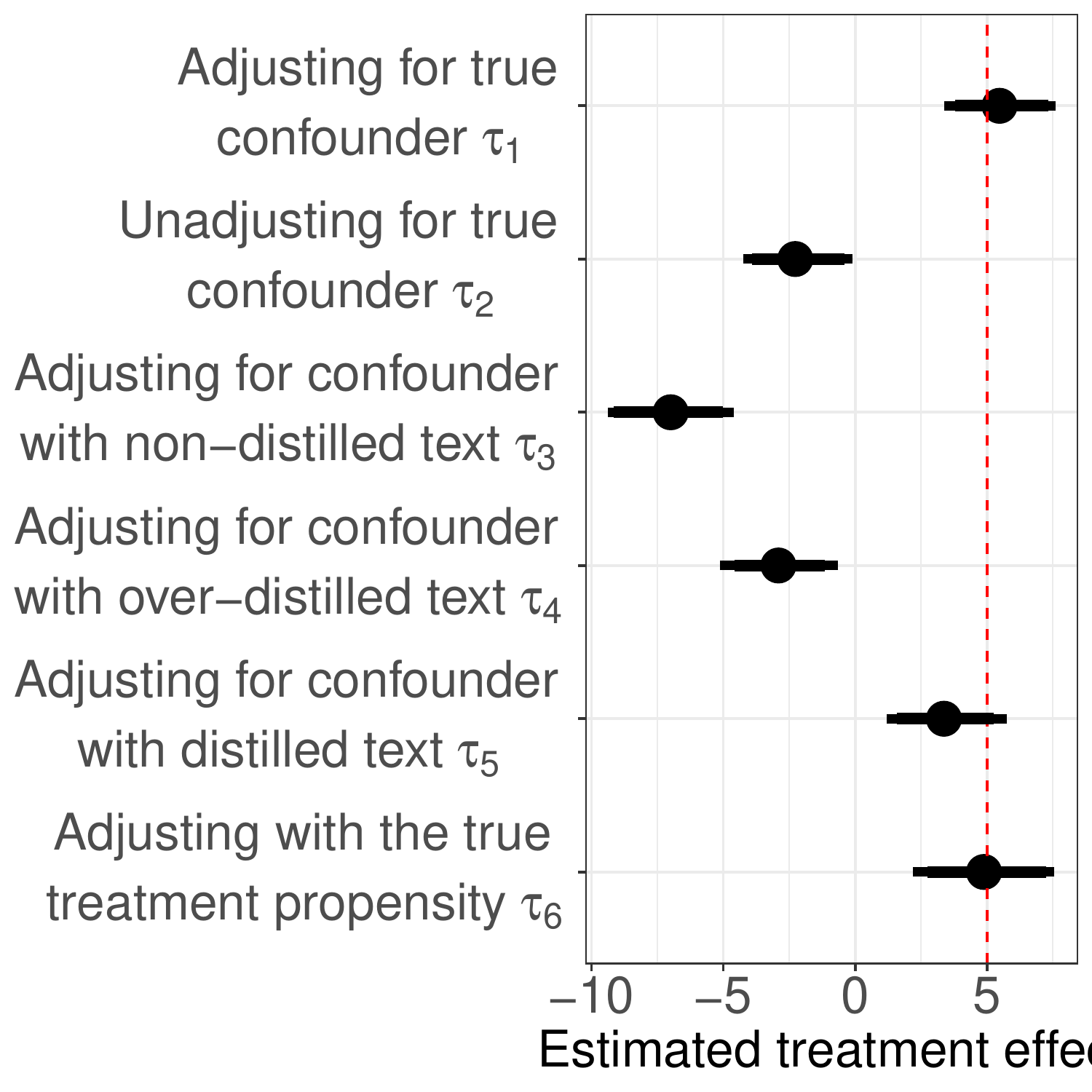}
\caption{Estimates under different distillation regimes. \label{fig:estimates}}
\end{center}
\end{figure}

Based on the setting described in \S\ref{sec:method}, our analysis produces six estimates, three based on distillation and three based on facts about the data-generating process. Figure~\ref{fig:estimates} shows all estimates.

\vspace{-0.75cm}

The \emph{\textbf{first estimate}}, $\hat{\tau}_1 = 5.5$, is the baseline where all information is known to the outcome model, including $U$. Because this linear model adjusting for $U$ and $\bX$ is equivalent to the data-generating model, and the estimated effect would be equal to the true value of 5 without sampling noise. The bootstrapped 95\% confidence interval (CI) is 3.4 to 7.6. The \emph{\textbf{second estimate}}, $\hat{\tau}_2$ = -2.3, is obtained when $U$ is omitted from the model to induce omitted variable bias (CI: -4.2, -0.1). 

The \emph{\textbf{third estimate}}, $\hat{\tau}_3$, 
uses IPW to estimate the ATE (see \S\ref{sec:ipwdetails}). Here, we use the non-disitilled documents, $\bW$, to estimate propensities.
As Figure 2 shows, in the absence of distillation, the bias \emph{increases} compared to conditioning on $\bX$ alone, producing $\hat{\tau_3}=-7.0$ (CI: -9.4, -4.6). 
The \emph{\textbf{fourth estimate}}, $\hat{\tau}_4$, applies overly aggressive distillation.  
This approach gives a result similar to the unadjusted estimate: $\hat{\tau_4}=-2.9$ (CI: -5.1, -0.6).

The \emph{\textbf{fifth estimate}}, $\hat{\tau}_5$, applies oracle distillation by removing the paragraphs we know were affected by $T$. 
Using $\bW^*$, the bias is reduced substantially, yielding an estimate $\hat{\tau_5}=3.5$ (CI: 1.2, 5.8). As the CI of this $\hat{\tau}$ includes the true $\tau=5$, we conclude that  
distillation successfully recovers $\tau$. However, we note that this recovery is not perfect and will be affected by sampling and modeling parameters. 


The \emph{\textbf{sixth estimate}}, $\hat{\tau}_6$, 
demonstrates the impact of model selection for the propensity estimator. 
Using the \emph{true} (simulated) propensity, the IPW estimate is $\hat{\tau_6}=4.9$ (CI: 2.2, 7.6). This result shows that further gains could be made by careful model selection 
\cite{chernozhukov2018double}.

\begin{figure}[htb]
 \begin{center}  \includegraphics[width=0.7\linewidth]{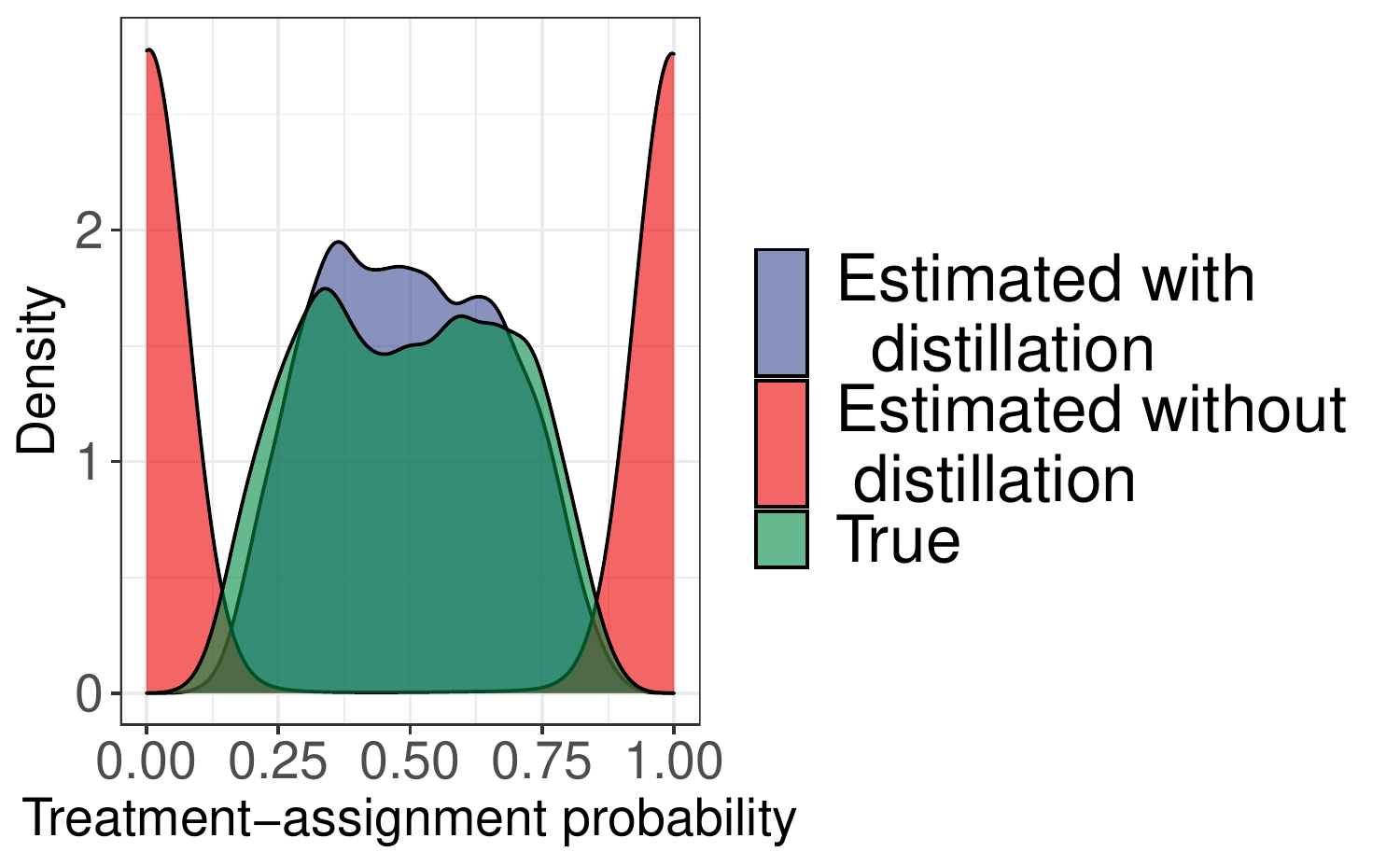}
\caption{Estimated and true assignment probabilities.
\label{fig:histogram}}
\end{center}
\end{figure}

Figure~\ref{fig:histogram} shows distributions of propensity values for $\hat{\tau}_3$, $\hat{\tau}_5$, and $\hat{\tau}_6$. 
Without distillation (red), the estimated propensities cluster near 0 and 1. $T$ is predicted almost perfectly, as mentioned in \S\ref{ss:case1}, causing the IPW estimate to to be similar to the unweighted one. Conversely, with distillation, the predicted probabilities are now similar to the data-generating propensities, and thereby, the resulting causal estimate is improved.


\vspace{-0.6cm}
\section{Discussion}
This paper shows the critical role of the no-treatment-leakage assumption when using text for causal inference. While text is becoming an established data source, it may harbour valuable information about a confounder but also contaminating information about post-treatment effects. 
%
This issue has seen little discussion in text-based causal inference literature \citep{mozer2020,roberts2020,feder2021,daoudStatisticalModelingThree2020}, but has the potential to severely bias causal estimates, potentially leading to false discoveries or invalid policy recommendations in social and health settings \cite{kinoScopingReviewUse2021,daoudImpactInternationalMonetary2017,balgiPersonalizedPublicPolicy2022}.

Before discussing the implication of treatment leakage, three limitations should be considered. First, more work is required to show how the no-treatment-leakage assumption operates under different covariance structures (i.e., different data-generating processes). Second, a larger simulation framework is needed to decompose estimator bias and variance. Third, all results are based on simulated data, and more research is needed to generalize our insights to real data. Although simulate data are idealized, they provide a benefit by allowing us to analyze the mechanics of treatment leakage and text distillation in a controlled environment. 
Based on our simulated data, our analysis shows that when the no-treatment-leakage assumption is violated,  effect estimates will be severely biased. In the presence of treatment leakage, scholars may be better off abstaining from using a non-distilled text to adjust for confounding. Although, in theory, the  best  solution is to use a text distillation that removes all treatment leakage, in practice, using distillation can be difficult to achieve. 

Therefore, one critical extension of our work is to develop methods that estimates the amount of treatment leakage in text. This estimate will enable applied researchers to make an informed decision about whether to adjusting for text-based confounding or abstain from it when leakage is high, and when text distillation is not an option.  

A second extensions is to develop a generalized framework that accounts for when the adjusted text represents multiple nodes in a DAG (in combination with the confounding and the treatment or without them). While our article focuses on treatment leakage, there are other types of leakage when a single document is a function of combinations of DAG nodes such as the outcome, confounder, treatment, mediator, or instrument. Thus, a generalization of the no-treatment-leakage assumption is the \emph{no-node-leakage assumption}. Such methods will benefit from insights established in the literature on causal inference with proxies \cite{penaMonotonicityNondifferentiallyMismeasured2020,vanderweelePrinciplesConfounderSelection2019,miao2018identifying,rissanenCriticalLookConsistency2022}. 
A third extension is to develop a variety of text distillation methods, suitable for different application settings. Researchers need alternative frameworks when human partitioning of text is not possible to achieve manually, because of corpus size or language complexities. \emph{Automatic} distillation could be attempted with additional assumptions, perhaps building from the literature on removing sensitive information in text representations \citep{bolukbasi2016,ravfogel2020}.



\section*{Acknowledgements}

Richard Johansson was supported by the projects \emph{Interpreting and Grounding Pre-trained Representations for NLP} and \emph{Representation Learning for Conversational AI}, both funded by Wallenberg  AI,  Autonomous  Systems  and  Software Program (WASP) funded  by  the  Knut  and  Alice Wallenberg Foundation. Adel Daoud would like to acknowledge a grant from The Royal Swedish Academy of Letters, History and Antiquities.

\bibliography{bibliography}
\bibliographystyle{acl_natbib}

\appendix

\section{Synthetic Data Generation}
\label{app:general}

We first summarize the general approach in this section and provide details for the simulation in \S\ref{sec:experiments} in the next section.

For each document $i$, we first draw observed and unobserved confounders $\bm{X}_i$ and $U_i$, and then the treatment $T_i$. For each paragraph $j$ in the document, we draw a paragraph topic $Z_{ij}$, depending on the values of $U_i$ and $T_i$, and then a 
prompt $W^0_{ij}$ depending on the value of $Z_{ij}$.
Finally, we sample from the GPT-2 language model\footnote{We used the implementation from the HuggingFace repository, \url{https://huggingface.co/gpt2}.} to generate the paragraph text $\bm{W}_{ij}$, starting from the prompt $W^0_{ij}$ and with a vocabulary distribution shift defined by $Z_{ij}$.
Algorithm~\ref{alg:syngen} shows the pseudocode.

\begin{algorithm}
\caption{Generation of synthetic data.\label{alg:syngen}}
\begin{tabbing}
\hspace{1mm}\textbf{for} $i \in 1, \ldots, N$\\
\hspace{5mm}$\bm{X}_i \sim f_X$\\
\hspace{5mm}$U_i \sim f_U$\\
\hspace{5mm}$T_i \sim \text{Bernoulli}(\text{sigmoid}(f_T(\bm{X}_i, U_i)))$\\
\hspace{5mm}$Y_i \sim f_Y(\bm{X}_i, U_i, T_i)$\\
\hspace{5mm}\textbf{for} $j \in 1, \ldots, K$\\
\hspace{9mm}$Z_{ij} \sim \text{Categorical}(f_{Z}(U_i, T_i))$\\ 
\hspace{9mm}$W^0_{ij} \sim \text{Categorical}(f_{W^0}(Z_{ij}))$\\
\hspace{9mm}$\bm{W}_{ij} \sim \text{LM}(W^0_{ij}, Z_{ij})$
\end{tabbing}
\end{algorithm}

In the pseudocode above, the functions $f_X$, $f_U$, $f_T$, and $f_Y$ define the distributions of the observed confounders, unobserved confounder, treatment and outcome, respectively.
On the paragraph level, the function $f_Z$ defines a categorical distribution over paragraph topics, and $f_{W^0}$ a categorical distribution over prompts.

Similarly to \newcite{wooddoughty2021}, we use two mechanisms to condition the generation of a paragraph on a topic $Z$: a prompt and a vocabulary distribution shift.
The distribution shift is designed to promote a set of \emph{keywords} related to the topic and we implement it by multiplying the language model probabilities by a topic-specific vector $\theta_Z$ of scale factors:
\[
P'(w|\text{context}, Z) \propto P_{\text{LM}}(w|\text{context}) \cdot \theta_Z(w)
\]
\section{Parameterization Used in \S\ref{sec:experiments}}
\label{app:specific}

In \S\ref{sec:experiments}, we generated $N$ = 10,000 instances, each consisting of numerical values and a document.
We used the following distributions to generate the document-level variables: $f_X$ was a 3-dimensional isotropic Gaussian; $f_U$ was an even coin toss; $f_T$ was linear in $\bm{X}_i$ and $U_i$; $f_Y$ was Gaussian with a mean defined by a linear function of $\bm{X}_i$, $U_i$, and $T_i$ and a fixed standard deviation.

Each document consisted of $K$ = 20 paragraphs. For the paragraph generation, we defined five different topics: two corresponding to positive and negative treatment values; two corresponding to positive and negative values of the unobserved confounder; one general background topic that was unrelated to $U$ or $T$ (but conceptually thought of as controlled by other ``residual'' variables $R$).
For a document with given values of $U$ and $T$, we set the topic distribution $f_Z$ to select the $U$ topic with a probability of 0.2, the $T$ topic with a probability of 0.2, and the general topic with a probability of 0.6.

The generated texts were designed to simulate a hypothetical use case where the researchers want to investigate the effect of IMF programs on some country-level indicator \citep[cf.][]{daoud2019international}. The treatment variable $T$ represents the presence or absence of an IMF program; the unseen confounder $U$ represents the political situation of the country with respect to the IMF.
For each topic except the general topic, we define four different prompts: for instance, for a positive treatment value, one of the prompts was \emph{The International Monetary Fund mandates the deregulation of [COUNTRY]'s labor market}. In the analysis, \emph{``[COUNTRY]''} is substituted by randomly sampled country names.

All topics except the general topic defined a distribution shift used when generating from the language model. We used 8 topic keywords for each of these topics. For these keywords, the corresponding entries in the vocabulary distribution shift vector $\log \theta_Z$ were set to a value that defines the strength of the effect of $T$ on $\bW$; for all other words except these keywords, $\log \theta_Z$ was 0. 
Since our focus in this paper is on a clear-cut use case where the effects are strong, we set the strength parameter to a value of 4, which gives a noticeable effect on the generated texts.

The text generation model was run on a single GPU (NVIDIA GeForce GTX TITAN X). Generating the 10,000 documents took around 10 hours.
The generation of random text is within the intended use of the GPT-2 model.

The implementation of the algorithm to generate the synthetic data is available in our repository.\footnote{\url{https://github.com/adeldaoud/AIforTextandCausalInference}}

\section{IPW Details}\label{sec:ipwdetails}
\subsection{Background}\label{ss:cifortext}
The ATE is defined as $\tau = \mathbb{E}[Y_i(1) - Y_i(0)]$, where $Y_i(t)$ is the potential outcome for unit $i$ under treatment $t$. It can be identified in randomized experiments \cite{Rubin1974}. However, the situation is  more complicated in the observational setting, where the treatment is not randomized to units but could be correlated with confounders, $\bX_i$, that are associated with the treatment and the outcome. In that setting, we can, with additional assumptions, still recover the ATE using Inverse Propensity Weighting (IPW) or related robust methods \cite{funk2011doubly}, where observations are weighted by the inverse of their estimated treatment probabilities $\hat{\pi}(\bX_i)=\widehat{\textrm{Pr}}(T_i=1|\bX_i)$ \cite{rosenbaum1983}: $\widehat{\tau} =  \frac{1}{n}\sum_{i=1}^n \left\{ \frac{T_i Y_i}{\hat\pi(\bX_i)} -\frac{(1-T_i)Y_i}{1-\hat\pi(\bX_i)}\right\}.$ 

\subsection{Estimation}
ATE estimates based on Inverse Propensity Weighting (see \S\ref{ss:cifortext}) 
require the estimation of the propensity scores, $\widehat{\Pr}(T|\bX, \bW)$. To estimate these scores, we applied a $L_1$-regularized logistic regression model using the \texttt{glmnet} package in \texttt{R}. The regularization strength ($\lambda$) was set automatically via 10-fold cross-validation. When estimating propensities, we represented the (non-distilled or distilled) document as an $L_2$-normalized TF-IDF vector using the 256 most frequent terms in the vocabulary, while the numerical covariates $\bX$ were standardized.

\begin{removed}
\begin{itemize}\addtolength{\itemsep}{-0.5\baselineskip}
    \item[$\hat{\tau}_1$] Idealized scenario with observed $U$.
    \item[$\hat{\tau}_2$] Biased estimate based on $X$ only.
    \item[$\hat{\tau}_3$] Biased estimate based on $X$ and the text $W$ without distillation.
    \item[$\hat{\tau}_4$] Estimate based on $X$ and the text $W$ with an over-aggressive distillation that removes all information about $U$ and $T$.  \item[$\hat{\tau}_5$] Estimate based on $X$ and $W$ with an idealized distillation that perfectly preserves the information about $U$ but completely removes the effect of $T$.  
    \item[$\hat{\tau}_6$] Idealized scenario with gold-standard propensity scores $P(T|X, U)$.
\end{itemize}

\section{Alternative material}
\subsection{intro alternative II}

In many scientific disciplines, scholars want to
understand how an \emph{intervention} on a \emph{treatment} variable may potentially affect an \emph{outcome} variable that we are interested in. However, in many cases the only experimental material that can be accessed is \emph{observational} data where researchers have no control over treatment assignment and \emph{confounder} variables interact with the treatment as well as with the outcome. Causal inference methods for observational data [CITE] control for the confounders to estimate the effect of intervening on the treatment variable.

However, in many realistic situations we cannot fully observe all variables that influence the causal system, which will cause estimates of causal effects to be biased if we do not account for them. Recently, several methods have been proposed to use \emph{text} as a proxy for an unobserved confounder \cite{keith2020,penaMonotonicityNondifferentiallyMismeasured2020}.

The reliability of treatment effect estimates with a text proxy is related to the strength of association between the text and the unobserved confounder [CITE]. However, if the text is also affected by the treatment variables, such estimates will be biased.
\todo{explain why}.
We refer to this situation as \emph{treatment leakage}.
\todo{explain why this problem is likely to arise in practice?}

\subsection{ATE Estimation}


We used IPW (see \S\ref{ss:cifortext}) to estimate the ATE.
IPW requires the estimation of the propensity $\hat{\pi}(T|X, W^{*})$ and to estimate these scores, we applied a $L_1$-regularized logistic regression model using the \texttt{glmnet} package in \texttt{R}. 
When estimating propensities, we represented the text $W$ as an $L_2$-normalized TF-IDF vector using the 256 most frequent terms in the vocabulary, while the numerical covariates $X$ were standardized.

\subsection{Experimental Setup}

\todo{justify why we are considering these different idealized scenarios}

We compute ATE estimates under the following six regimes:
(1) Idealized scenario with observed $U$, (2) biased estimate based on $X$ only, (3) biased estimate based on $X$ and the text $W$ without distillation, (4) estimate based on $X$ and the text $W^{*}$ with an over-aggressive distillation that removes all information about $U$ and $T$, (5) estimate based on $X$ and $W^{*}$ with an idealized distillation that perfectly preserves the information about $U$ but completely removes the effect of $T$. (6) idealized scenario with gold-standard propensity scores $P(T|X, U)$.
These estimates are compared to the true value of $\tau=5$ that was used to generate the synthetic data.

\todo{I know that this overlaps with the description in sec 4, but I think it can be useful to be explicit about the experimental strategy}

We motivate our study by alluding to the IMF case. Yet the IMF case generalizes to wide-ranging non-experimental settings such as the effect of social media on mental health, or using medical records in health care. As text is a general data source, an increasing number of research domains face the problem of information leakage. While much remains to be done, by highlighting the problem of treatment leakage, our article contributes to improving the design of causal inference when text is involved.


Our proposal here is to include an extra pre-preprocessing step to analyses that use text in estimating treatment effects. In particular, we propose that researchers identify the portions of the text which contain the signature of $T$ and remove them so that the arrow from $T$ to $W$ in Figure \ref{fig:causalmodel} is severed. We denote the post-distilled text as $W^{*}$. The resulting analysis can then safely condition on $W^{*}$ without risking the violation of the overlap assumption. 

This approach is not without assumptions. In particular, we assume that the text available to researchers can, indeed, be decomposed into portions involving $T$ and $U$. We call this the \emph{Text Separability Assumption}:
\begin{align*} 
&\textrm{\emph{Text Separability: }} W = \{f(T), g(U)\}
\end{align*} 
for some functions $f$ and $g$. In other words, the text can be decomposed into a portion (e.g., a set of paragraphs) that is determined only by $U$ and another portion that is determined by the eventual treatment. This assumption would be violated if the treatment information affected the entire tone of the document, so that no feature of the text was unaffected by the treatment receipt. If this assumption holds, we can remove $f(T)$ from $W$ to form the distilled $W^{*}$, which we can safely condition on in our analysis of $T$.

\subsection{Characterizing Treatment Leakage}
\label{ss:bias}
We define \emph{treatment leakage} as when  the text, $\bW$, is affected by treatment status, $T$.
\[
\textrm{ \emph{Treatment leakage:} } \; \exists \; \bW_T, \bW_U \subseteq \bW, 
\]
so that there is a part of the text affected by $T$ (i.e., $\bW_T$) and a part affected by $U$  (i.e., $\bW_U$). In general, these two parts could be intersecting, if, for example, both the confounder and the treatment affect the overall tone (sentiment) of a document.


The degree of treatment leakage can vary. First, it depends on how much signal $T$ injects into $\bW_T$. Second, it also depends on the relative (distributional) strength of each partition $\bW_U$ and $\bW_T$ in jointly composing the entire document $\bW$. Additionally, there will be random noise $R$ irrelevant for $T$ and $U$, yet present in composing $\bW$. If we simplify and assume that each partition carries strength equal to the number of its elements (e.g., words) and each element has the same strengths, then the magnitude of treatment leakage is: $\frac{|\bW_T|}{|\bW_U \cup \bW_T \cup R|}$, taking values $(0,1)$.

\end{removed}

\end{document}